\documentclass[preprint]{elsarticle}

\usepackage{lineno,hyperref}

\usepackage{xcolor}

\usepackage{amssymb}

\usepackage{graphicx}

\usepackage{amsmath}
\usepackage{algorithmic}
\usepackage{array}
\usepackage{fixltx2e}

\usepackage{stfloats}
\usepackage{url}

\usepackage[font=footnotesize,labelfont=bf]{caption}
\usepackage[font=footnotesize,labelfont=bf]{subcaption}

\usepackage{picins}

\usepackage{multirow}
\modulolinenumbers[5]

\journal{Elsevier}

\begin{document}

\begin{frontmatter}

\title{Exploring DeshuffleGANs in Self-Supervised Generative Adversarial Networks \footnote{https://doi.org/10.1016/j.patcog.2021.108244}}


\author[mymainaddress]{Gulcin Baykal\corref{mycorrespondingauthor}}
\cortext[mycorrespondingauthor]{Corresponding author}
\ead{baykalg@itu.edu.tr}

\author[mymainaddress]{Furkan Ozcelik}
\ead{ozcelikfu@itu.edu.tr}

\author[mymainaddress]{Gozde Unal}
\ead{gozde.unal@itu.edu.tr}

\address[mymainaddress]{Istanbul Technical University, Istanbul, Turkey}

\begin{abstract}
Generative Adversarial Networks (GANs) have become the most used networks towards solving the problem of image generation. Self-supervised GANs are later proposed to avoid the catastrophic forgetting of the discriminator and to improve the image generation quality without needing the class labels. However, the generalizability of the self-supervision tasks on different GAN architectures is not studied before. To that end, we extensively analyze the contribution of a previously proposed self-supervision task, deshuffling of the DeshuffleGANs in the generalizability context. We assign the deshuffling task to two different GAN discriminators and study the effects of the task on both architectures. We extend the evaluations compared to the previously proposed DeshuffleGANs on various datasets. We show that the DeshuffleGAN obtains the best FID results for several datasets compared to the other self-supervised GANs. Furthermore, we compare the deshuffling with the rotation prediction that is firstly deployed to the GAN training and demonstrate that its contribution exceeds the rotation prediction. We design the conditional DeshuffleGAN called cDeshuffleGAN to evaluate the quality of the learnt representations. Lastly, we show the contribution of the self-supervision tasks to the GAN training on the loss landscape and present that the effects of these tasks may not be cooperative to the adversarial training in some settings. Our code can be found at \url{https://github.com/gulcinbaykal/DeshuffleGAN}.
\end{abstract}


\end{frontmatter}

\section{Introduction}

Generative Adversarial Networks (GANs) \cite{gan} followed and surpassed the pioneering Autoencoders \cite{autoencoder} in employing deep neural networks (DNNs) for image generation with a certain quality in mimicking a desired data distribution. The main objectives of the image generation problem can be summarized as: generating high-quality images; generating diverse images; and achieving training stability \cite{gansurvey}. Numerous GAN variants attempt at producing photo-realistic image generations while the latter are desired to vary and capture a considerable amount of different modes in the real data distribution.

The GAN model's ability to learn the structures and the semantics of the data improves the quality of the generations synthesized by the generator of the model. As the standard training mechanism of the GANs proposed in \cite{gan} needs additional guidance in order to support the generator for synthesizing high-quality images that capture the real data distribution, typically, the capacity of the discriminator part of the GAN is enhanced to provide this guidance. While the enhancement of the discriminator is attained by auxiliary signals and tasks, the increased representation differentiating capacity of the discriminator introduces an enriched representation of the real data.

Following the unconditional GAN model, conditional GAN models are proposed in conditional image generation where initially, the generations are conditioned on class labels. While some of the conditional GAN models make use of the auxiliary signals of the labels in training, some others' discriminators also try to predict these labels as the auxiliary task. Benefiting from the auxiliary signals and tasks can help to optimize the training of the GANs while it supports the generator to produce high-quality images.

Considering that the enhancement of the discriminator with the auxiliary tasks improve the training stability and image generation quality of the GANs, the choice of the auxiliary task deserves additional attention. Since the auxiliary signals are attained by the labels of the datasets, GAN training with the unlabeled datasets becomes a critical problem. Most of the datasets used as benchmarks in GANs are unlabeled, and labeling them according to a criterion requires time and human resources which are expensive to have. Organization of datasets with labels is also not always sensible while there are already some datasets satisfying the essential properties of being large and diverse other than having labels. As a remedy, a recent framework, known as the \textit{self-supervised learning} can be utilized for compounding self-supervision tasks on unlabeled datasets to the GAN training in order to enhance the capacity of the discriminator at no computation and human resource cost.

To deploy the self-supervised learning into GANs, Chen et al. \cite{ssgan} propose the first self-supervised GAN, the SS-GAN, with the auxiliary task of predicting the rotation angles of the input images for the discriminator. Later, Tran et al. \cite{ssgan_minimax} enhance the self-supervised task of rotation predictions by a multi-class minimax game. Huang et al. \cite{fxgan} propose an auxiliary task for the discriminator to predict whether the input image is unmodified or not. The modification is the feature exchanging on the feature maps attained from the hidden layer of the discriminator. In DeshuffleGAN\footnote{A preliminary version of our work is presented at IEEE International Conference on Image Processing 2020 \cite{deshufflegan}.}, Baykal et al. \cite{deshufflegan} present that taking advantage of jigsaw puzzle solving as an auxiliary task for the discriminator improves the structural consistency in the synthesized images by the generator.

Despite introduction of various self-supervised GANs, there are still relevant concerns and questions regarding the proposed methods. In \cite{ssgan}, although the addition of the rotation prediction task is shown to improve the performance of the GAN training, the task should be also evaluated with the other self-supervision tasks in a comparative manner. As Tran et al. \cite{ssgan_minimax} evaluate their method on lower resolutions datasets, a more challenging problem of generating higher resolution images should be the primary concern. Furthermore, while Huang et al. \cite{fxgan} enhance the discriminator to predict whether the pixel exchange is performed or not, predicting the order of the pixel block exchanges as Baykal et al. \cite{deshufflegan} propose would lead to learning high-level features since the model is forced towards extracting these high-level features in order to capture the relative relationships of the patches. A crucial point to evaluate should be the quality of the learnt features since one of the important assertions of the self-supervised GANs is that self-supervision aids learning better representations while alleviating the catastrophic forgetting problem \cite{catastrophic} of the discriminator. To that end, Chen et al. \cite{ssgan} prove the contribution of their method to solve the aforementioned problem whereas Tran et al. \cite{ssgan_minimax} and Huang et al. \cite{fxgan} do not evaluate the quality of the features that are learned by the self-supervised GANs compared to those of the baseline methods with no self-supervision. Additionally, the  self-supervision tasks are not evaluated using different network architectures in Chen et al. \cite{ssgan} and Huang et al. \cite{fxgan}. Although different architectures are used in Tran et al. \cite{ssgan_minimax}, the effects of the architecture on the performance are not analyzed. Ultimately, we raise new questions about the contribution of learning trajectories for solving the self-supervision tasks on the generations, and the interaction of different networks with the self-supervision tasks.

All these concerns and questions lead us to extend the DeshuffleGANs \cite{deshufflegan} in a way that the contribution of the deshuffling task to the natural looking, structurally coherent image generation should be presented using various datasets. The representations learnt by the cooperation of the deshuffling should be compared to the representations learnt by the baseline model. The deshuffling and the other self-supervision tasks in the literature should be studied not only with the generation quality perspective, but also by considering their own learning curves. To do so, the loss landscapes of the self-supervision tasks could be analyzed from a visual stability aspect. Furthermore, the effects of the deshuffling task on various network architectures could be also observed and analyzed.

In this work, we present and extend the self-supervised GAN model, the DeshuffleGAN, that utilizes deshuffling task of jigsaw images in order to train a discriminator that learns high-level features through resolving the spatial configuration of the input data for generating structurally consistent images. Our contributions can be listed as follows:

\begin{itemize}
    \item We compare the deshuffling task with the other self-supervision tasks deployed to the discriminator on various datasets that are mostly used in GAN benchmarks as LSUN-Bedroom \cite{lsun}, LSUN-Church \cite{lsun} and CelebA-HQ \cite{progan} in terms of generation quality.
    \item We study the effects of deshuffling auxiliary task on training through 2 different networks unconditionally. The first one consists of ResNet blocks as Miyato et al. \cite{spectral} propose while the second one is designed based on DCGAN \cite{dcgan} architecture as Jolicoeur-Martineau proposes in \cite{ragan} which makes use of relativistic adversarial training. We analyze the results considering the network architectures.
    \item We design the cDeshuffleGAN, the conditional version of the original model, and showcase the conditioned generation quality of the cDeshuffleGAN visually.
    \item We evaluate the representation quality of the features learnt by the cDeshuffleGAN on ImageNet \cite{imagenet} dataset.
    \item We study the effects of the self-supervision tasks on the loss landscape.
\end{itemize}

DeshuffleGAN surpasses the other self-supervision tasks in terms of generation quality which is evaluated with the Fr$\acute{e}$chet Inception Distance (FID) \cite{fid} measure whose reasonableness is empirically shown by Lucic et al. \cite{equal_gan}. It improves the representation quality compared to the baseline architecture which is evaluated as Zhang et al. \cite{colorization} proposed.

\section{Related Work}

\subsection{Generative Adversarial Networks}

After the idea of two adversarial players is introduced by Goodfellow et al. \cite{gan}, GANs received accelerating attention for the task of image generation. In order to satisfy the objectives of the image generation, various types of GANs are designed.

Some studies try to overcome the problem of unconditional training instability by enforcing the discriminator to satisfy the Lipschitz condition \cite{gp}. Lipschitz condition is sought to be satisfied by adding a spectral weight regularization \cite{spectral} or spectral bounding using efficient computations \cite{spectral_bounding}. Some other studies change the loss definitions of standard GAN with the Wasserstein loss \cite{wasserstein}, the least squares loss \cite{lsgan} or the  relativistic loss \cite{ragan}. Another study combines multiple generators to learn different aspects of the data via orthogonal vectors \cite{tackle}. These studies also aim to overcome the vanishing gradient problem of the generator or the mode collapse problem.

Aside from the unconditional GANs, conditional GANs are proposed to make use of the class labels of the data as the auxiliary signals. Labels for the input images are typically represented as vectors. Odena et al. \cite{cgan} utilize the label vectors directly while Miyato et al. \cite{projection} project the label vectors instead of concatenation. The incorporation of the auxiliary signals into the adversarial training improves the stability and the quality of the class conditional image generations. In InfoGAN \cite{infogan} and ACGAN \cite{acgan}, the discriminator tries to predict the class labels as the auxiliary task which enhances the representation capacity of the discriminator.

Various architectural designs are also proposed during the development of both unconditional and conditional GANs. DCGAN \cite{dcgan} proposes the usage of convolutions in GANs which results in the generation of higher resolution images. Progressive GANs \cite{progan} extend the network architecture progressively where the network includes millions of parameters and improves the generation quality while the training strategy improves stability. Self-Attention GANs \cite{sagan} deploy the usage of attention mechanisms in order to learn global dependencies for generating images and to capture mode diversity with the help of a large receptive field. \cite{sagan} is able to produce higher resolution images by deploying the spectral normalization to both the discriminator and the generator networks while ensuring to remedy the problem of vanishing gradients. BigGANs \cite{biggan} introduce a deeper architecture that is based on the architecture of \cite{sagan} and proposes to use very large batch sizes which require to have an enormous GPU power. \cite{biggan} also deploys the attention mechanism and spectral normalization in order to obtain diversity in generations and to overcome the training stability problem.

\subsection{Self-Supervised Learning for GANs}

Self-supervised learning is a two-staged method which is a subset of unsupervised learning and designed in order to learn qualified feature representations via training on unlabeled datasets. The first stage of the self-supervised learning is to use unlabeled datasets and label them with \textit{pseudo-labels} which are produced automatically by manipulating the data in a way that the manipulation is not expensive besides its result is known already. Data with the pseudo-labels are used subsequently in a task called \textit{pretext} which is designed for the DNN to solve with the purpose of learning visual features that encode high-level semantic information of the data. The second stage of self-supervised learning is to use the learnt visual features in other computer vision tasks called \textit{downstream} tasks where the downstream tasks can be treated as the evaluation methods for the quality of the learnt features. The term ``quality'' indicates the representability and the generalizability of the features which is obviously desired for an efficient use in high-level problems.

The mostly used image-based pretext tasks can be categorized as \textit{generation-based methods} which learn the features while generating images compatible with a data distribution, and \textit{context-based methods} which learn the features while trying to capture context similarities or spatial context structures \cite{sssurvey}. In generation-based methods such as Zhang et al. \cite{colorization} where the task is the colorization of grey-scale images, the discriminator network learns the semantic features of the images and the learnt features by the discriminator can be beneficial for the downstream tasks. Another example of generation-based methods is image inpainting where the task is the completion of the missing regions in the images. Wang et al. \cite{laplacian} used Laplacian pyramid-based GANs for face completion which can be given as an example of an inpainting task. In context-based methods, a network is trained for a pretext task that can be designed for various context-related problems. Noroozi et al. \cite{jigsaw} present a self-supervision task where the task is the prediction of shuffling order of the input image that is divided into 9 tiles like a jigsaw puzzle and shuffled with a random permutation which leads to learning spatial context structures of the images. Gidaris et al. \cite{rotation} present the task of predicting the rotation angle of the input image where the input image might be rotated by one of 4 different angles and the learning process gives results to obtain semantic information of the data such as object structures. Doersch et al. \cite{context} introduce the task of finding the relationship between the positions of the two image patches which is managed by learning the high-level semantic information of the data. 

Some of these classical pretext tasks are also deployed to the GAN training. Chen et al. \cite{ssgan} manage to obtain training stability by the enhancement of the discriminator's representation power in a way that overcomes the catastrophic forgetting problem to a certain degree \cite{catastrophic} with the cooperation of the rotation prediction task. While Tran et al. \cite{ssgan_minimax} also use the rotation prediction task, Huang et al. \cite{fxgan} propose the usage of a context-based method as a pretext task for the discriminator where the task is simply to predict whether there is a feature exchange on the feature maps or not. Baykal et al. \cite{deshufflegan} use jigsaw puzzle solving pretext task in their DeshuffleGAN.

Beyond the design of the pretext tasks which is basically critical for the self-supervised GANs, notable questions are also answered such as the effects of the DNN choices for different pretext tasks by Kolesnikov et al. \cite{revisitingss}, the effects of employing multiple pretext tasks by Doersch et al. \cite{multitaskss}, and the usage of multiple-domain data by Feng et al. \cite{multidomainss}. All these works present studies of self-supervised learning methods towards a deeper understanding of the workings of the idea of self-supervision, which in turn aids in understanding of the interaction between the GAN training and the self-supervised learning framework.

\section{Methodology}

\begin{figure*}[t]
    \centering
    \includegraphics[width=\linewidth]{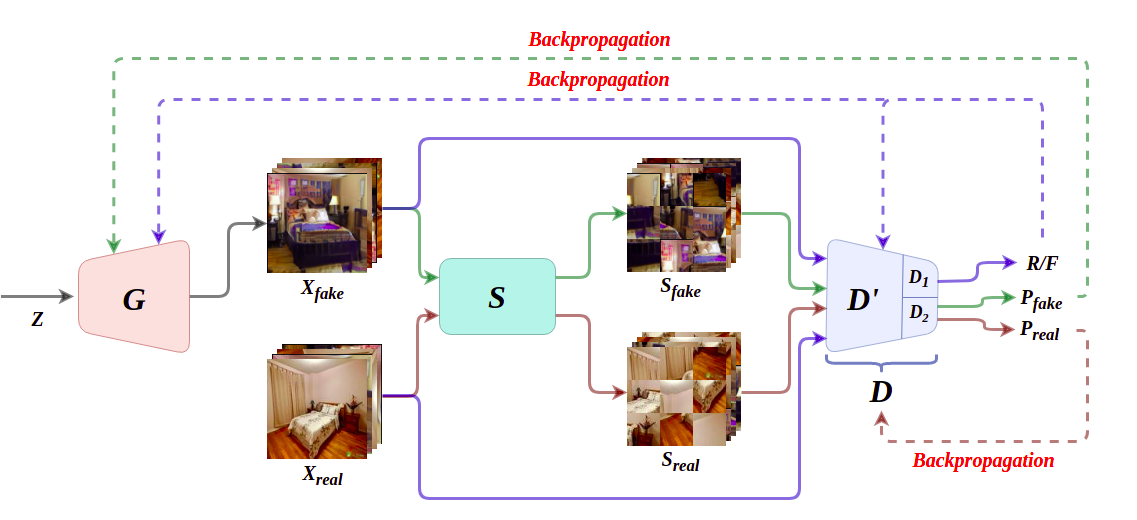}
    \caption{Proposed training mechanism for DeshuffleGAN. $X_{fake}$ represents a batch of generated images while $X_{real}$ represents the real images. $\textit{S}$ is the Shuffler and it shuffles $X_{fake}$ and $X_{real}$ as $S_{fake}$ and $S_{real}$ respectively. $P_{fake}$ and $P_{real}$ are the permutation predictions for $S_{real}$ and $S_{fake}$. Green and Red lines indicate the shuffling and deshuffling operations, including their backpropagation paths. Purple lines refer to the standard adversarial training of the GAN. $D'$ represents the shared layers of the deshuffling and the real/fake prediction tasks. $D_{1}$ and $D_{2}$ are the output layers for the real/fake prediction and the deshuffling tasks, respectively.}
    \label{fig:main_figure}
\end{figure*}

The proposed training mechanism for DeshuffleGAN is shown in Fig.~\ref{fig:main_figure}. The standard adversarial training is visualized along with the auxiliary task training. 

In the adversarial training, generator $G$ takes a vector $z$ which is sampled from the Gaussian normal distribution as $z \sim N(0, 1)$ with the size of $128 \times 1$ as input and learns to decode the latent space to the image space while the synthesized samples are named as $X_{fake}$. As the discriminator $D$ learns to distinguish between the real samples and the fake samples, samples derived from the real data distribution are named as $X_{real}$ and included to the adversarial training of $D$ along with $X_{fake}$. While $D$ learns the discriminative features of the real data, it gives feedback to $G$ that helps $G$ in generating images compatible with the real data distribution. In classical adversarial training, the only feedback given to $G$ from $D$ is the prediction of whether the samples are real or fake.

In the training for the auxiliary self-supervision task, the main objective is to enrich the learnt features by $D$ in a way to have well-defined structures that lead to the generation of structurally coherent and consistent images by $G$. To arrange the input data with pseudo-labels, a new component that is light in computation, which is obviously desired, is designed as the $S$ block in Figure~\ref{fig:main_figure}, denoting the $Shuffler$ operation. 

$S$ is designed to divide its image inputs into 9 square tiles with the size of \textit{n'/3} and shuffle them with an order of permutation. $n$ is the resized input image size and it may not be divisible by 3. Then, the biggest number smaller than $n$ and divisible by 3 is obtained as $n'$ and used to calculate the tile sizes. The generations in this work are $128 \times 128$ sized, so $n'$ is used as 126, and each image tile is obtained as a square with the size of $42 \times 42$. The $n' \times n'$ sized top left area of the input image is used to construct a jigsaw puzzle of $3 \times 3$ in the shuffling process of $S$. 

As there are 9 tiles, there are 9! possible permutations for shuffling, which is impractical to use. The number of permutations as in \cite{jigen} is set to 30. The permutations are selected according to the Hamming distances proposed in \cite{jigsaw}. Each sample in the input batch of $S$ is shuffled by a different permutation out of 30, and the permutations applied to the input batch are selected randomly in order to prevent memorization. Since both $X_{real}$ and $X_{fake}$ play roles in the auxiliary task, they are shuffled by $S$ to produce $S_{real}$ and $S_{fake}$, respectively. To obtain the same input size $n$, $S_{real}$ and $S_{fake}$ are padded by 1 pixel on each side with a replication padding.

Originally, $D$ has an output layer as the last stage of real/fake (r/f) prediction, which we name as $D_{1}$. Since the cooperation of the auxiliary task needs an architectural enhancement on $D$, one additional output layer that learns to map the output of $D'$ to permutation indices is added as $D_{2}$. In DeshuffleGAN, $D'$ has the $same$ shared weights for both of the tasks while $D_{1}$ layer outputs the r/f predictions for $X_{real}$ and $X_{fake}$. $D_{2}$ layer outputs the permutation indices for $S_{real}$ and $S_{fake}$. According to architectural design, output layers can be convolutional or linear.

DeshuffleGAN's $D$ is trained to capture the structural features within the input data by the deshuffling auxiliary task so that it can encourage $G$ to generate structurally coherent images as $D$ can deshuffle the shuffled versions of the generated images. As $D$ is meant to learn structural features, it can easily deshuffle $S_{fake}$ if $X_{fake}$ is structurally consistent and includes meaningful pieces so that the puzzle tiles in $S_{fake}$ are related to each other and obtain a continuity when they are deshuffled. On the contrary case of image generations that do not contain meaningfully structured elements, $D$ signals $G$ in a way to support it to synthesize well-structured and correctly composed images so that it can manage deshuffling $S_{fake}$.

The objectives for both adversarial and auxiliary training become more of an issue in that the choices for the objectives affect the performance of the model, therefore they need to be studied in detail.

\subsection{Adversarial Objective}

The standard GAN training proposed in \cite{gan} has distinct objectives for $D$ and $G$, defined as:

\begin{align}
    L_{\theta} &= -\mathbb{E}_{x_r \sim P} [\log(D'(x_r))] - \mathbb{E}_{x_f \sim Q} [\log(1 - D'(x_f))] \nonumber\\
    L_{\phi} &= -\mathbb{E}_{x_f \sim Q} [\log(D'(x_f))] \nonumber\\
    D'(x_r) &= sigmoid(D(x_r))  \nonumber\\
    D'(x_f) &= sigmoid(D(x_f))  \label{eq:1} 
\end{align}
where $x_r$ is the real data that is sampled from the real data distribution $P$, $x_f$ is the generated data that is sampled from generated data distribution $Q$. $\theta$ and $\phi$ are the parameters of the $D$ and $G$ network models, respectively. $D(x)$ is the non-transformed output of $D$ that measures the realness of the input $x$. In classical GAN training,  cross-entropy loss is used as the error measure in Eq.~\eqref{eq:1}. Sigmoid is the nonlinear transformation over the output of $D$ in \cite{gan}. $L_\theta$ and $L_\phi$ refer to the objective functions for $D$ and $G$, respectively.

Mao et al. \cite{lsgan} propose to disuse sigmoid as a transformation for the output of $D$ in order to mitigate the vanishing gradient problem and modifies the adversarial objectives as:

\begin{align}
    L_{\theta} &= \frac{1}{2}\mathbb{E}_{x_r \sim P} [(D(x_r) - 1)^2] + \frac{1}{2} \mathbb{E}_{x_f \sim Q} [D(x_f)^2] \nonumber\\
    L_{\phi} &= \frac{1}{2}\mathbb{E}_{x_f \sim Q} [(D(x_f) - 1)^2]. \label{eq:2} 
\end{align}

Later, Miyato et al. \cite{spectral} deploy the hinge loss to their GAN training as:

\begin{align}
    L_\theta = &- \mathbb{E}_{x_r \sim P} [min(0, -1 + D(x_r))] \nonumber\\
    &- \mathbb{E}_{x_f \sim Q} [min(0, -1 - D(x_f))] \nonumber\\
    L_\phi = &- \mathbb{E}_{x_f \sim Q} [D(x_f)]. \label{eq:3} 
\end{align}

Even if the formulations change according to different concerns such as stability and optimality, in all of the training mechanisms shown with Eq. ~\eqref{eq:1} - \eqref{eq:3}, $D$ returns a probabilistic result for the input being real as $G$ tries to boost the probability of the fake data being real. Relativistic GANs \cite{ragan} argue that $D$ should converge to 0.5 in order to maximize the JS-divergence between $X_{real}$ and $X_{fake}$ while the classical adversarial training actually causes $D$ to output 1 for all data. It is also shown that when $D$ is optimal, its gradients mostly come from the fake samples. Therefore, $D$ later cannot learn from the real images. As a solution, the ``relativism'' idea is introduced in \cite{ragan}, as the aim of the GAN training should be not only to increase the probability that the fake data are real, but also to decrease the probability that the real data are real. That means the realness of the input is not considered separately but considered relatively. Then the relativism idea is further enhanced to define a more global $D$ as the author says, in a way that the average $D(x)$ results for the random samples of opposing type are used in the calculations and the objectives integrate ``relativistic average'' idea.

The relativistic average objective used in this work deploys the relativism idea into the least squares objective formulized in Eq.~\eqref{eq:2} as:

\begin{align}
    L_\theta &= \frac{1}{2}\mathbb{E}_{x_r \sim P} [(D(x_r) - \mathbb{E}_{x_f \sim Q}D(x_f) - 1)^2] \nonumber\\
    &+ \frac{1}{2}\mathbb{E}_{x_f \sim Q} [(D(x_f) - \mathbb{E}_{x_r \sim P}D(x_r))^2] \nonumber\\
    L_\phi &= \frac{1}{2}\mathbb{E}_{x_r \sim P} [(D(x_r) - \mathbb{E}_{x_f \sim Q}D(x_f) + 1)^2] \nonumber\\
    &+ \frac{1}{2}\mathbb{E}_{x_f \sim Q} [(D(x_f) - \mathbb{E}_{x_r \sim P}D(x_r) - 1)^2] \label{eq:4} 
\end{align}

As numerous GAN models exist, we evaluate the performances of two different GAN architectures as the backbone in our self-supervised DeshuffleGAN. We utilize the hinge loss objective formulated in Eq.~\eqref{eq:3} for the modified GAN architecture proposed in \cite{spectral}, named as SNGAN. We utilize the relativistic average least squares objective (RaLSGAN) formulated in Eq.~\eqref{eq:4} for the DCGAN \cite{dcgan} architecture as in \cite{ragan}. We add one more linear layer to the discriminator of the SNGAN architecture, and one convolutional layer to the discriminator of the DCGAN architecture as $D_2$ for the deshuffling task as in Figure~\ref{fig:main_figure}.

As the adversarial training concerns only $X_{real}$ and $X_{fake}$ for the r/f predictions, $S_{real}$ and $S_{fake}$ are not included in the adversarial objectives.

\subsection{The Deshuffler Objective}

Besides the adversarial objectives, DeshuffleGAN involves additional objectives for $D$ and $G$ in order to increase the learning capability of $D$ by deshuffling the shuffled images in order to improve the quality of the generation that is produced by $G$.

For $D$, the objective is to correctly predict the shuffling order of $S_{real}$, meaning that the difference between $P_{real}$ (permutation predictions for $S_{real}$) and the true random permutations used by $S$ to shuffle $X_{real}$ should be minimized. As $D$ should learn solely the features of the real data, it should be updated in a way to minimize the error calculated only considering $S_{real}$. Cooperation of $S_{fake}$ to the optimization of $D$ would lead to a problem of learning fake and unnecessary features of the generated data hence $G$ would not be supported to synthesize images that are compatible to the real data distribution. The cross-entropy loss is used as the deshuffler optimization objective for $D$:

\begin{equation}
    V_\theta = -\sum_n^N y_d^n \ln \tilde{y}_d^n
    \label{eq:5}
\end{equation}
where N denotes the number of samples in an input batch, $y_d^n$ is the one-hot encoded label vector of size 30x1 applied to the $n^{th}$ sample of $X_{real}$ and $\tilde{y}_d^n$ is the $n^{th}$ permutation prediction vector of $P_{real}$. $\theta$ are the parameters of the $D$ network.

For $G$, the objective is to generate high quality images $X_{fake}$ for which $D$ can correctly predict the shuffling order of $S_{fake}$. Therefore, the error between $P_{fake}$ and the random permutations used by $S$ to shuffle $X_{fake}$ is desired to be a minimum. As $D$ learns the features of the real data, if $S_{fake}$ can be deshuffled by $D$, $G$ can be expected to generate images compatible to the real data distribution. On the other hand, if the generations are not structurally consistent and do not capture the features of the real data, then $D$ which learns the features of the real data cannot deshuffle $S_{fake}$, thus sends a signal to $G$ to synthesize qualified images so that it can deshuffle $S_{fake}$. The cross-entropy loss is used as the deshuffler optimization objective for $G$:

\begin{equation}
    V_\phi = -\sum_n^N y_g^n \ln \tilde{y}_g^n
    \label{eq:6}
\end{equation}
where N denotes the number of samples in an input batch, $y_g^n$ is the one-hot encoded label vector of size 30x1 applied to the $n^{th}$ sample of $X_{fake}$ and $\tilde{y}_g^n$ is the $n^{th}$ permutation prediction vector of $P_{fake}$. $\phi$ are the parameters of the $G$ network.

\subsection{Full Objective of the DeshuffleGAN}

We formulate the full objective as the combination of the adversarial objective and the deshuffler objective as follows:

\begin{equation}
    \begin{split}
        L_\theta^D = L_\theta + \alpha V_\theta \\
        L_\phi^G = L_\phi + \beta V_\phi
    \end{split}     \label{eq:DeshuffleGANLoss}
\end{equation}
where $\alpha$ and $\beta$ parameters determine the influence of the objectives in the overall training. In our experiments, $\alpha$ and $\beta$ are used as 1 and 0.5, respectively. As we follow the same procedure in \cite{ssgan}, we directly choose $\alpha$ to be 1, while we select $\beta$ according to the experimental results, where $\beta=0.5$ provided the best performance in terms of the FID measure, which is described in Section~\ref{sec:generationquality}.

The training of the DeshuffleGAN model involves optimization of the objective functions $L_\theta^D$ and $L_\phi^G$ with respect to $\theta$ and $\phi$ parameters, respectively for the $D$ and the $G$ networks as follows:

\begin{equation}
    \begin{split}
        \theta = \arg\min L_\theta^D \\
        \phi = \arg\min L_\phi^G.
    \end{split}     \label{eq:argmin}
\end{equation}

\begin{figure}[t]
    \centering
    \includegraphics[width=\linewidth]{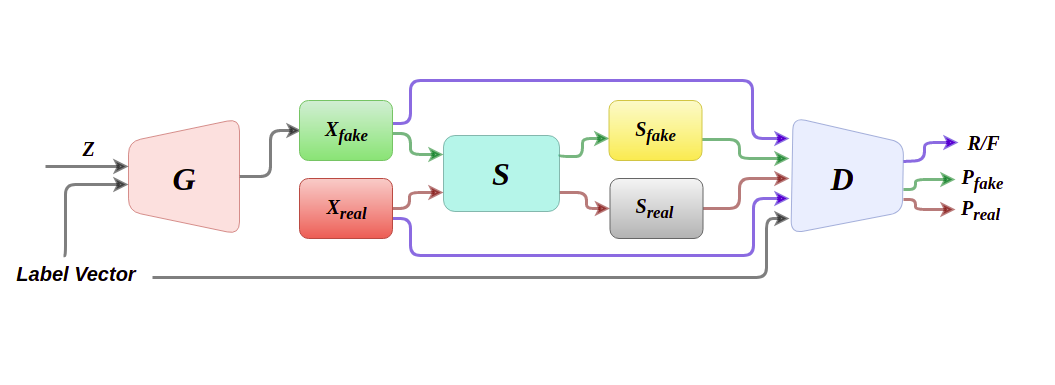}
    \caption{Proposed model for cDeshuffleGAN. The training mechanism is the same as the DeshuffleGAN while the cDeshuffleGAN is also supported with the class label.}
    \label{fig:cgan}
\end{figure}

\subsection{cDeshuffleGAN}

As an extension to the proposed DeshuffleGAN, we also train a conditional DeshuffleGAN, which we call cDeshuffleGAN. Fig.~\ref{fig:cgan} shows the proposed model of the cDeshuffleGAN. We utilize the conditional SNGAN in \cite{spectral}, cSNGAN, as the baseline architecture for cDeshuffleGAN. The main difference between the DeshuffleGAN and the cDeshuffleGAN is the utilization of the label vector in the training. The main training objective for the cDeshuffleGAN is to evaluate the representation quality learnt by $D$ on the ImageNet dataset. Therefore, we enhance the capacity of the SNGAN which is relatively a smaller network with not only the auxiliary self-supervision signal but also with the class label in order to obtain meaningful generated samples of ImageNet. By doing so, we perform quantitative evaluations of representation quality learnt by the cSNGAN and cDeshuffleGAN.

\section{Experiments}

\subsection{Experimental Settings}

In our work, we show the effects of deshuffling self-supervision on GAN training using various datasets and GAN architectures.

As the datasets, we choose LSUN-Bedroom, CelebA-HQ, and LSUN-Church to compare the self-supervision tasks deployed to GANs and to analyze the effects of the self-supervision on GANs in terms of losses. We also use ImageNet in order to evaluate the representation quality of the features.

We utilize two distinct adversarial objectives for different backbones: (i) we experiment with the relativistic loss on the RaLSGAN which is based on the DCGAN \cite{dcgan} architecture; (ii) we experiment with hinge-loss on the modified version of the SNGAN architecture where we use the proposed channel sizes by Kurach et al. \cite{channel_info} for the $D$ and the $G$. As the reduced channel sizes were empirically reported to improve the performance in \cite{channel_info}, we also utilize these small differences which simplify the architecture design in our implementation. For ImageNet experiments, we employ the conditional SNGAN as the baseline and deploy deshuffling unit over its architecture. We use the same modified channel sizes for the conditional SNGAN and modify the $G$ and $D$ architectures for conditioning as in \cite{spectral}.

We use Adam optimizer \cite{adam} for the RaLSGAN $D$ and $G$ with (0.5, 0.999) parameters, and for the SNGAN $D$ and $G$ with (0, 0.9) parameters, respectively. Learning rate is set to 0.0002, and batch size is set to 64 for all experiments. We choose the number of total iterations as 200K for the LSUN-Bedroom, CelebA-HQ, and LSUN-Church while we perform 1M iterations on ImageNet. We select the number of $D$ iterations per $G$ as 1 and 2 for the RaLSGAN, and the SNGAN, respectively. We conduct all of the experiments on a single V100 GPU. We use the PyTorch framework for the implementations.

\begin{figure}[t]
    \begin{subfigure}{0.48\textwidth}
        \includegraphics[width=\linewidth]{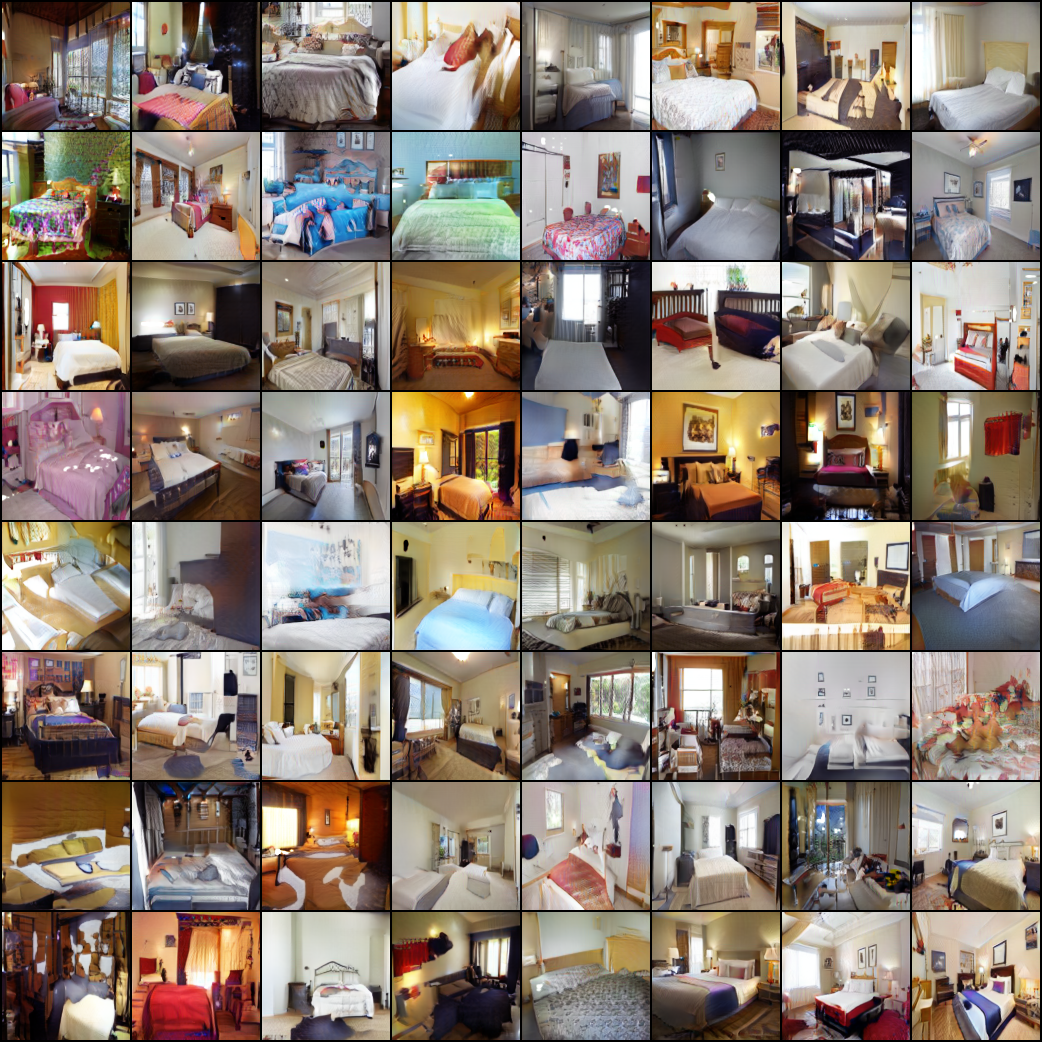}
        \caption{LSUN-Bedroom}
    \end{subfigure}
    \hspace*{\fill}
    \begin{subfigure}{0.48\textwidth}
        \includegraphics[width=\linewidth]{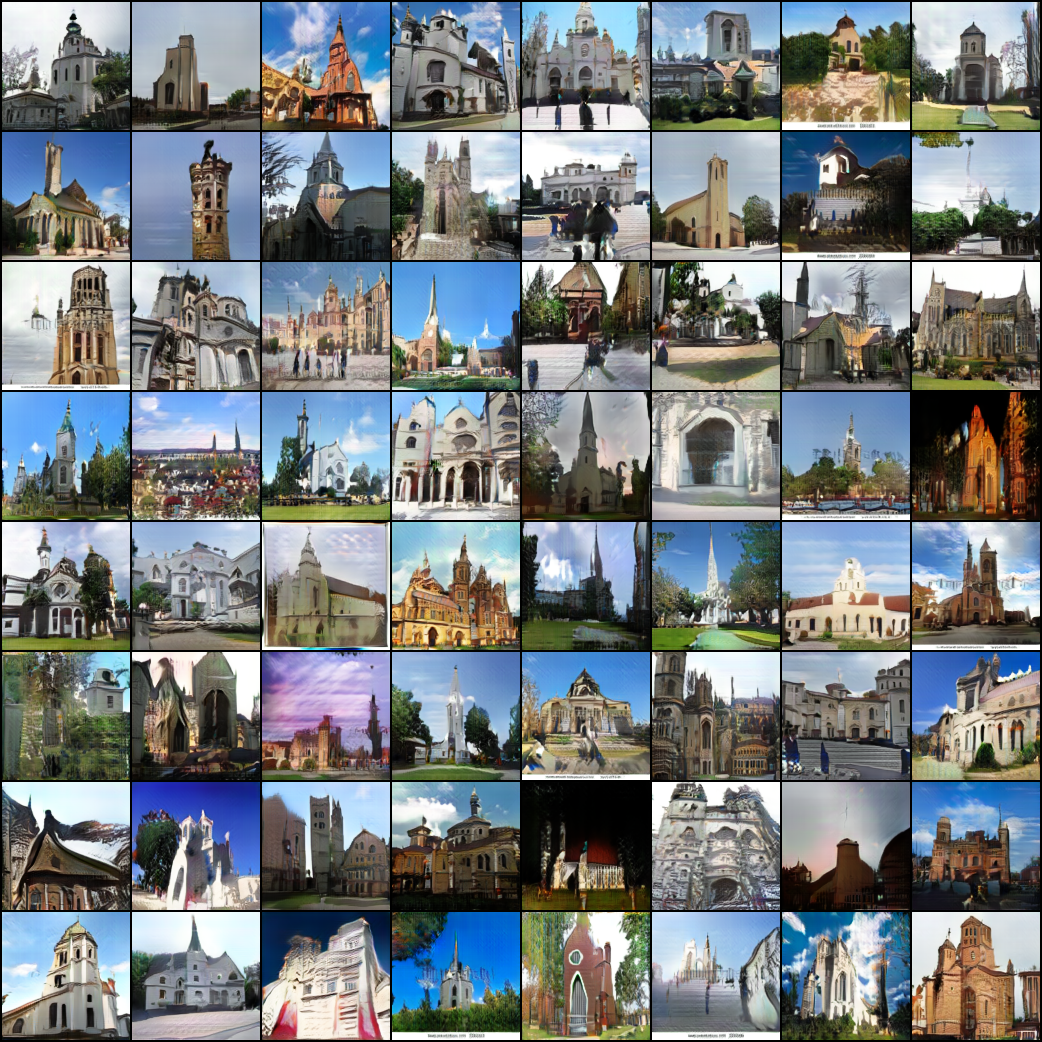}
        \caption{LSUN-Church}
    \end{subfigure}
    \caption{Randomly generated samples for (a) LSUN-Bedroom and (b) LSUN-Church by the best-performing DeshuffleGANs.} 
    \label{fig:examples}
\end{figure}

\begin{figure}[t]
    \begin{subfigure}{0.31\textwidth}
        \includegraphics[width=\linewidth]{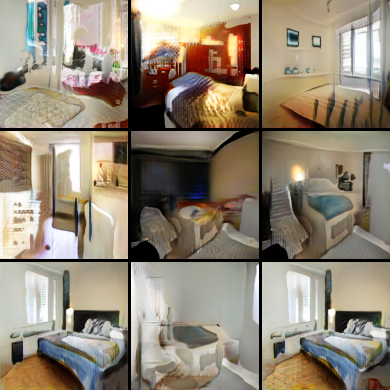}
        \caption{SNGAN}
    \end{subfigure}
    \hspace*{\fill}
    \begin{subfigure}{0.31\textwidth}
        \includegraphics[width=\linewidth]{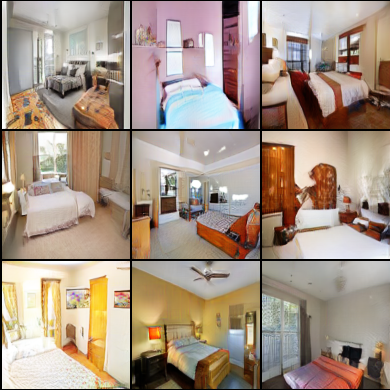}
        \caption{SNGAN + Deshuffle}
    \end{subfigure}
    \hspace*{\fill}
    \begin{subfigure}{0.31\textwidth}
        \includegraphics[width=\linewidth]{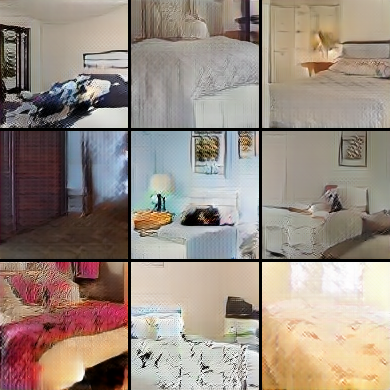}
        \caption{SS-GAN}
    \end{subfigure}
    
    \medskip
    \begin{subfigure}{0.31\textwidth}
        \includegraphics[width=\linewidth]{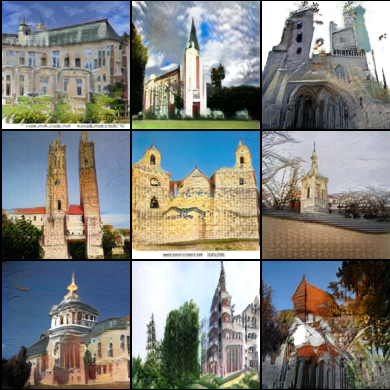}
        \caption{SNGAN}
    \end{subfigure}
    \hspace*{\fill}
    \begin{subfigure}{0.31\textwidth}
        \includegraphics[width=\linewidth]{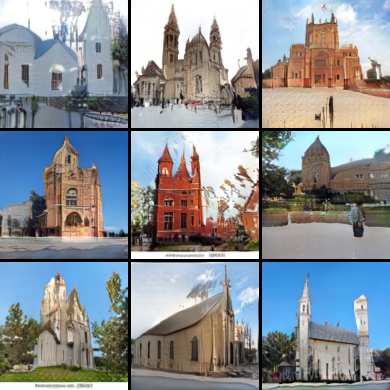}
        \caption{SNGAN + Deshuffle}
    \end{subfigure}
    \hspace*{\fill}
    \begin{subfigure}{0.31\textwidth}
        \includegraphics[width=\linewidth]{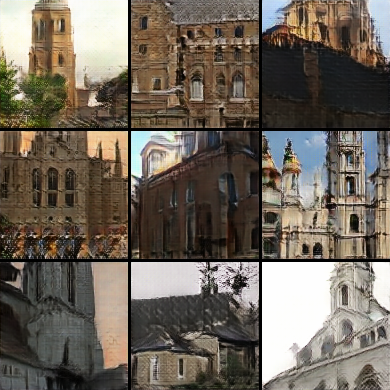}
        \caption{SS-GAN}
    \end{subfigure}
    
    \caption{Randomly generated samples for (a)-(c) LSUN-Bedroom and (d)-(f) LSUN-Church datasets by the best-performing Generators of the corresponding GANs.} 
    \label{fig:compare_examples}
\end{figure}

\begin{table}[h]
  \begin{center}
    \caption{Best FID results obtained for each dataset.}
    \label{tab:eval}
    \begin{tabular}{ccc} 
        \hline
        DATASET &METHOD &FID \\
        \hline
        \multirow{6}{*}{LSUN-Church} & SS-GAN &33.82 \\
        &FX-GAN &-\\
        &SNGAN &22.17\\
        &\textbf{SNGAN + Deshuffle} &\textbf{14.01}\\
        &RaLSGAN &22.02\\
        &RaLSGAN + Deshuffle &21.47\\
        \hline
        \multirow{6}{*}{LSUN-Bedroom} & SS-GAN \cite{ssgan} &13.30\\
        &FX-GAN \cite{fxgan} &12.90\\
        &SNGAN &51.22\\
        &\textbf{SNGAN + Deshuffle} &\textbf{10.13}\\
        &RaLSGAN &34.76\\
        &RaLSGAN + Deshuffle &36.48\\
        \hline
        \multirow{6}{*}{CelebA-HQ} & SS-GAN \cite{ssgan} &24.36\\
        &\textbf{FX-GAN} \cite{fxgan} &\textbf{19.25}\\
        &SNGAN &20.55\\
        &SNGAN + Deshuffle &32.46\\
        &RaLSGAN &25.29\\
        &RaLSGAN + Deshuffle &22.51\\
    \end{tabular}
  \end{center}
\end{table}

\subsection{Evaluations}

\subsubsection{Generation Quality}
\label{sec:generationquality}

\begin{figure*}[t]
    \centering
    \includegraphics[width=\linewidth]{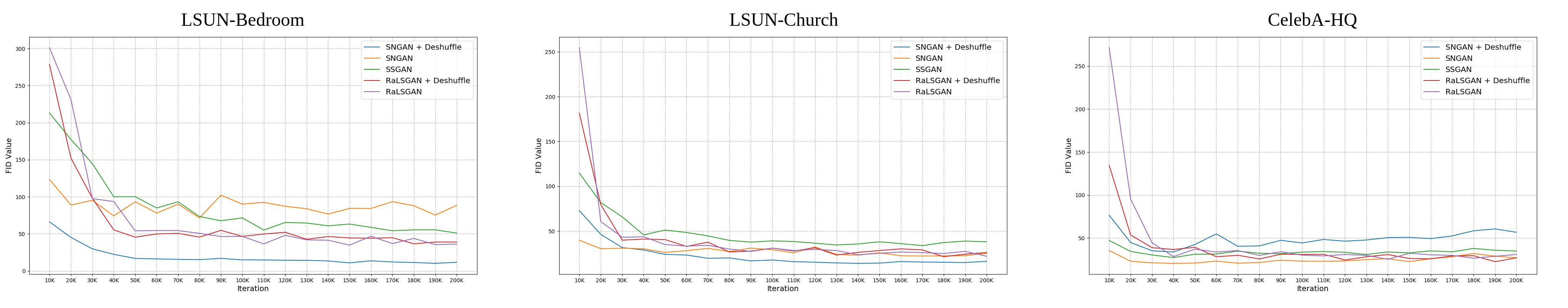}
    \caption{FID curves for each of the datasets. For LSUN datasets, the DeshuffleGANs achieve the best performances and the training stability of the DeshuffleGANs reflects the smoothness of the FID curves.}
    \label{fig:fid_all}
\end{figure*}

Visual results synthesized by the best-performing DeshuffleGANs both for the LSUN-Bedroom and the LSUN-Church datasets are shown in Fig~\ref{fig:examples}. Aside from displaying the visual performance of the DeshuffleGANs on its own, Fig~\ref{fig:compare_examples} shows the visual comparison between the baseline, the DeshuffleGAN, and the SS-GAN results. In order to numerically evaluate the generation quality of the DeshuffleGANs on different datasets, we use the most generally preferred evaluation metric FID \cite{fid}, and compare our method with the other methods. FID compares the statistical metrics of the real data distribution and the generated data distribution where the lower difference between the measurements indicate more similarity between the distributions, which is obviously the desired case. A comparison between the methods on the datasets are given in Table~\ref{tab:eval}. From the literature, we report the best FID values that are obtained for the LSUN-Bedroom and CelebA-HQ datasets for the SS-GAN from \cite{ssgan} and the FX-GAN from \cite{fxgan}. For the LSUN-Church dataset, we run the official implementation of \cite{ssgan}. As the official implementation of the FX-GAN \cite{fxgan} is not published, we do not report the values of the FX-GAN for the LSUN-Church dataset. FID curves of all the methods compared in Table~\ref{tab:eval} are shown in Fig~\ref{fig:fid_all}.

\begin{figure}[t]
    \centering
    \includegraphics[width=0.8\linewidth]{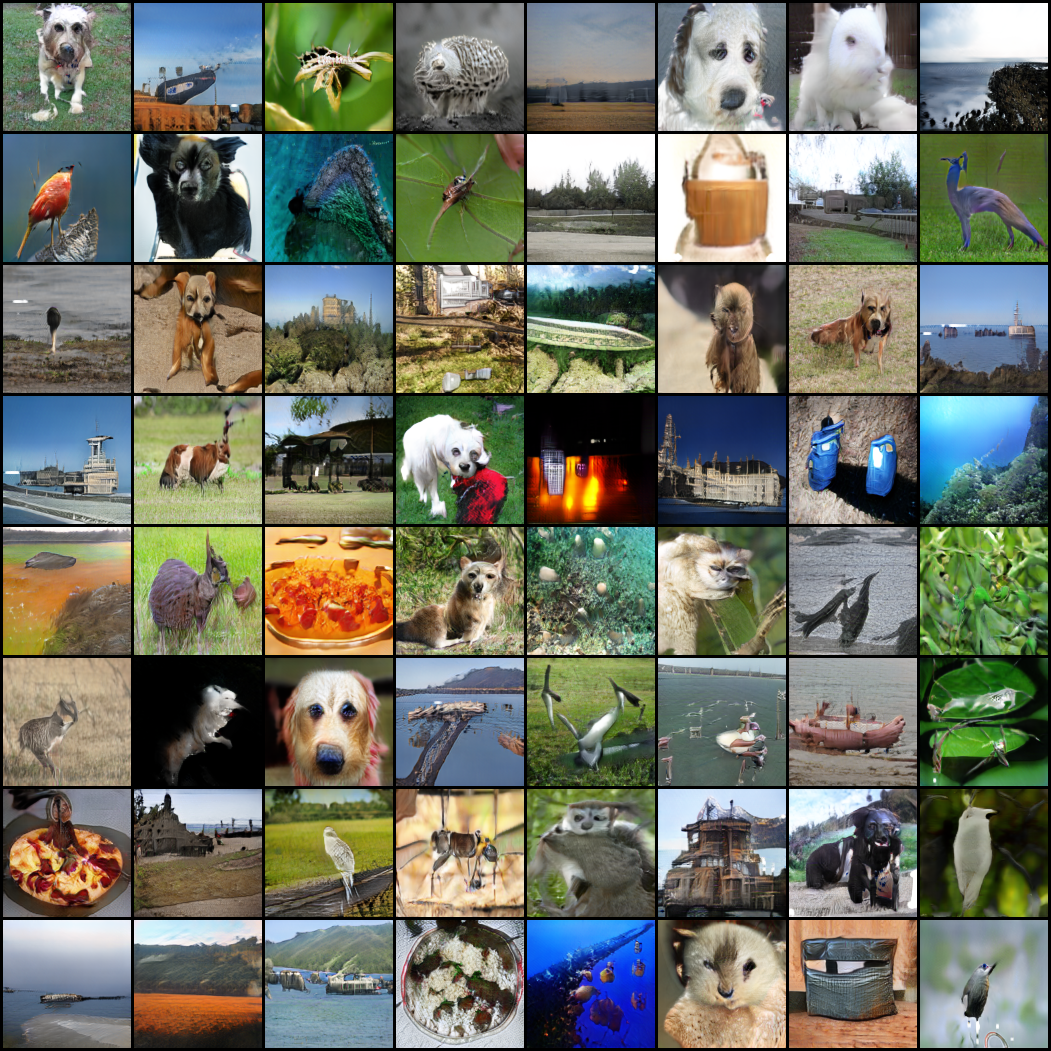}
    \caption{Randomly generated samples for ImageNet by the conditional DeshuffleGAN.}
    \label{fig:imagenet_examples}
\end{figure}

In Fig~\ref{fig:imagenet_examples}, we present class-conditional generations of ImageNet by the conditional DeshuffleGAN. The motivation for the ImageNet training is to evaluate the quality of the representations by a commonly used method where the main objective is not to generate high-quality images conditionally which is only manageable with massive networks and enormous computational power. Nonetheless, we manage to generate visually substantial samples with a relatively smaller network and computational power than the successful networks like BigGANs \cite{biggan}.

\subsubsection{Representation Quality}

In order to test whether the learnt representations of the $D$ are meaningful or not, we use a commonly preferred evaluation method proposed in \cite{colorization}, which basically suggests training a linear layer on the representations that are extracted from the hidden layers of $D$ and use the trained layer for the classification task on ImageNet. For the representation quality evaluation, we use conditional SNGAN. The $D$ of the SNGAN consists of 6 ResNet blocks and we extract the representations of the last block which is the $6^{th}$ block as they represent the high-level features.

\begin{figure}[t]
    \centering
    \includegraphics[width=\linewidth]{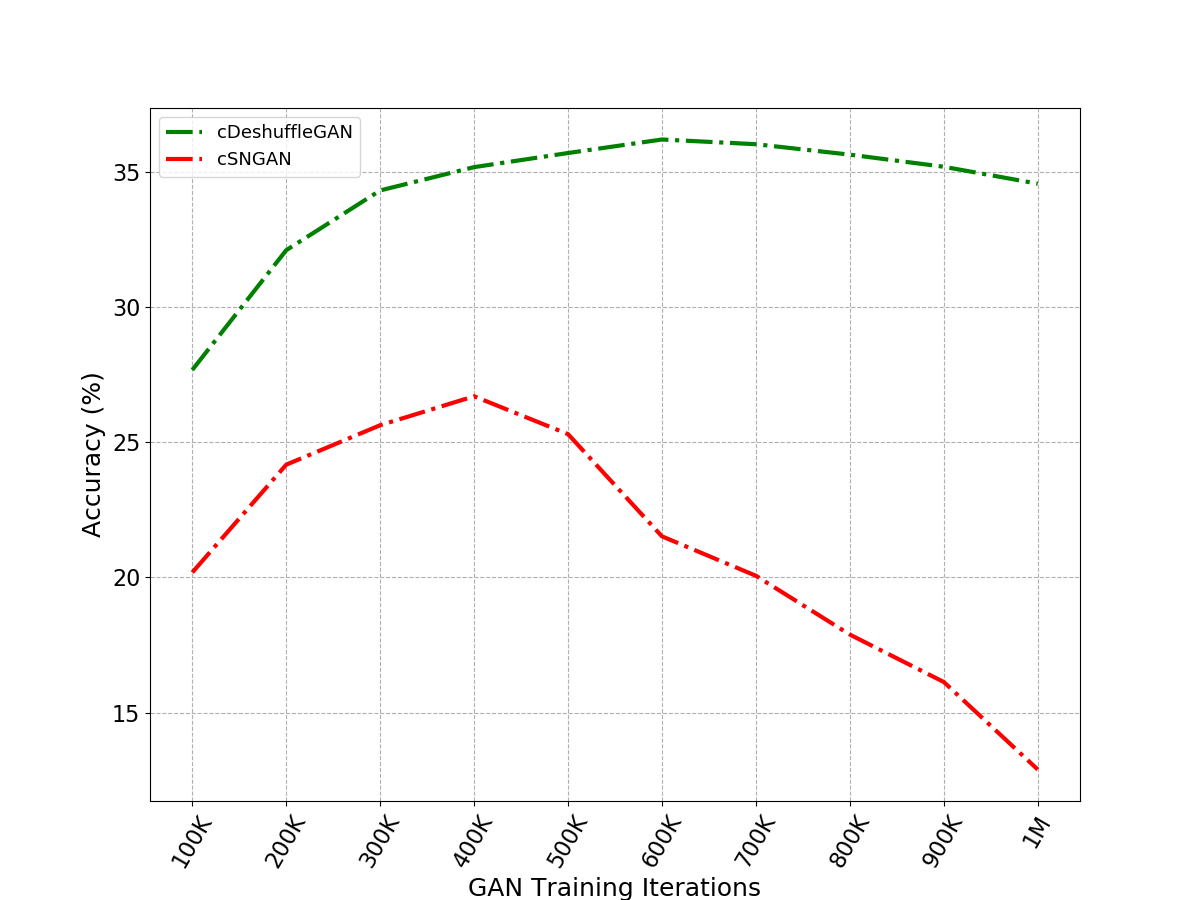}
    \caption{Top-1 accuracies of the regressors trained on the final layer representations of the discriminators attained at each 100K iterations.}
    \label{fig:imagenet}
\end{figure}

In Fig.~\ref{fig:imagenet}, we show that the linear layer trained on the representations extracted from the conditional DeshuffleGAN performs better than the linear layer trained on the representations extracted from the original conditional SNGAN on ImageNet test data. These results provide evidence to our hypothesis that the learnt features of the DeshuffleGAN model increases the effectiveness of the representation power of a baseline GAN model, here exemplified by the original SNGAN.

\begin{figure*}[t]
    \centering
    \includegraphics[width=0.99\linewidth]{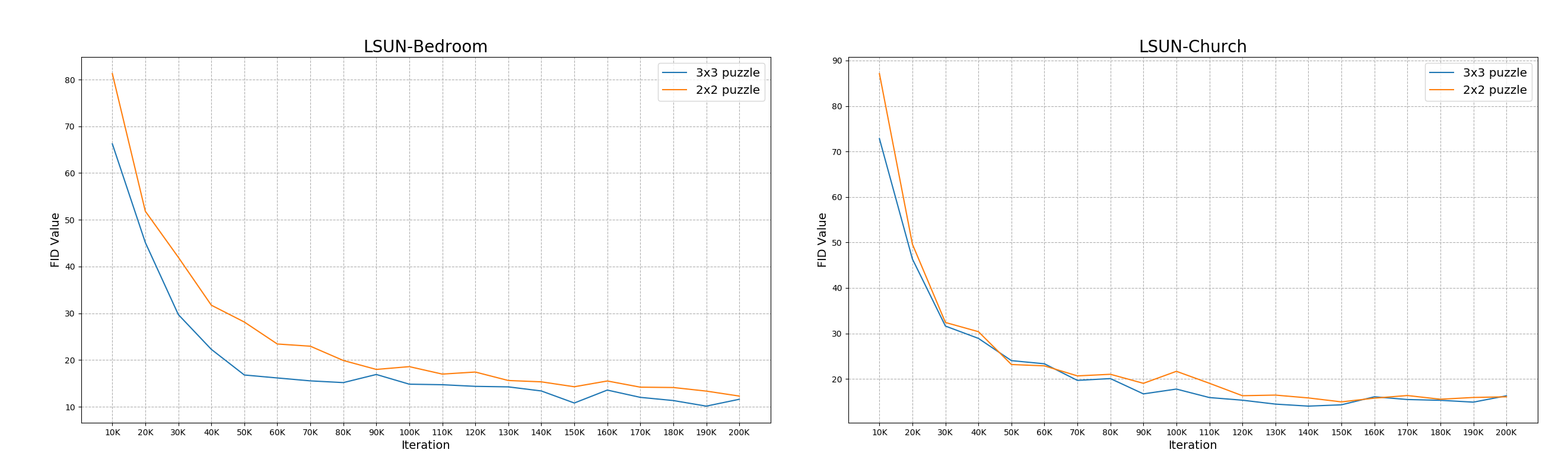}
    \caption{FID curves of the DeshuffleGANs whose discriminators are trained with the objective of solving $2 \times 2$ and $3 \times 3$ jigsaw puzzles on LSUN datasets.}
    \label{fig:perm_comp}
\end{figure*}

\subsubsection{Deshuffling Performance}

In order to finalize the design of the aforementioned Shuffler operation $S$, we conduct some experiments using 4 tiles and 9 tiles. We start our experiments with 9 tiles as the $3 \times 3$ puzzle configuration is studied in the pioneering work by Noroozi et al. \cite{jigsaw}. We then utilize the 30 permutations set proposed in \cite{jigen} since its practical usage is shown. We hypothetically suppose that solving a $3 \times 3$ puzzle would be more complicated than solving a $2 \times 2$ puzzle, and this challenge would lead to learn better representations. In order to evaluate our supposition experimentally, we generate all the possible 24 permutations of a $2 \times 2$ puzzle, and train our DeshuffleGAN with the objective of deshuffling a $2 \times 2$ puzzle on the LSUN datasets. Fig.~\ref{fig:perm_comp} shows that solving a $3 \times 3$ puzzle yields a better generation performance in terms of FID for the LSUN datasets which supports our assumptions.

As the number of permutations is the decisive factor in terms of time complexity, the cost of solving a $2 \times 2$ puzzle is similar to solving a $3 \times 3$ puzzle in our case because the number of permutations for both 9 tiles and 4 tiles are quite similar. We observe that the training time for the baseline SNGAN architecture is $\approx$ 2.5 hours per 10K iterations on a single V100 GPU while the training time for the DeshuffleGAN is $\approx$ 3.5 hours per 10K iterations. As the training is an offline process and the deshuffling does not affect the inference time, the amount of difference in the training can be affordable.

Other than the choice of the number of the tiles and the time complexity of the deshuffling operations, the accuracy of the permutation order prediction actually matters since the whole idea is based on the assumption that a discriminator can learn to deshuffle. To evaluate the performance of the discriminator for the deshuffling task, we utilize the DeshuffleGAN with the SNGAN backbone. The trained discriminator serves the purpose to deshuffle both real and fake samples, therefore, we train the discriminator on the training data split of a given dataset, and test it for classification of the deshuffling order over both the real and fake samples of the data. We also test a discriminator that is trained on a given dataset to deshuffle samples for another dataset to observe the generalizability of the learnt features. Table~\ref{tab:accuracy} reports those performance results for different train and test data sets. The high accuracy results of the discriminator in terms of the deshuffling order prediction demonstrate that the key conjecture behind the DeshuffleGAN is satisfied.

\subsection{Discussion}

We show that the modified architecture of the SNGAN with the addition of deshuffling task outperforms other methods and achieves better performances on LSUN-Bedroom and LSUN-Church datasets in terms of both numerical and visual results as the Table~\ref{tab:eval},  Fig.~\ref{fig:examples}, and Fig.~\ref{fig:compare_examples} indicate, respectively. We present that while the deshuffling operation yields an FID improvement by 80.2\% compared to the poorly performing SNGAN baseline over LSUN-Bedroom, it yields an improvement by 36.8\% compared to the SNGAN baseline over LSUN-Church.

While Fig.~\ref{fig:examples} depicts the generation quality and the diversity of the DeshuffleGANs on LSUN datasets, Fig.~\ref{fig:compare_examples} visually demonstrates the superior performance of the deshuffling task over the rotation task, and the baseline setting. We observe that the baseline SNGAN suffers from the mode collapse problem and the objects are not structurally consistent in the LSUN-Bedroom dataset. Similar issues arise for the SNGAN over the LSUN-Church dataset where the buildings seem structurally deformed even if it can capture different modes from the dataset.

The generation results with the official SS-GAN implementation shows that the rotation task worsens the generation performance compared to that of the baseline over the LSUN-Bedroom dataset. The objects in the images seem noisy and they cannot represent a general bedroom scene. For the LSUN-Church dataset, SS-GAN generates images quite similar to each other. SS-GAN fails to generate reasonable quality in the synthesized images even if some images show structural consistency.

Compared to the SNGAN and the SS-GAN results on both LSUN-Bedroom and LSUN-Church datasets, SNGAN + Deshuffle setting improves the generation quality, as it can be observed that it can synthesize diverse, natural, and structurally coherent objects and scenes. We especially monitor the effects of the deshuffling on the objects in the indoor scenes, and the church buildings in the outdoor scenes as they have originally structural integrity, which the deshuffling task helps to maintain in the synthesized images.

Likewise the visual results, we also demonstrate the superiority of the deshuffling task over the other self-supervision pretext tasks in Table~\ref{tab:eval} in terms of the FID measure. We motivate our work initially with the intuition that jigsaw puzzle solving is a more complex task than the rotation prediction and the feature exchange detection tasks, we also validate the benefits of deshuffling against other self-supervision pretext tasks experimentally. While the rotation prediction task is a global transformation problem, a task like deshuffling which is apparently manageable only if the structural context is learnt by the network, requires better representations at both global and local levels.  We show both visual and numerical evidence to the statement that the deshuffling task is more complicated than the rotation prediction task through our exploration of self-supervised GANs with those tasks. We further show that the feature exchange detection which is similar to deshuffling yet not as much as a complex task cannot beat the performance of the deshuffling on LSUN-Bedroom datasets in terms of FID. Although the FX-GAN \cite{fxgan} seems to be the best performing method on the CelebA-HQ dataset, the deshuffling task can also improve the performance of the baseline RaLSGAN architecture on the CelebA-HQ dataset. The reported differences may happen due to dataset characteristics as well as the architectural design of the GAN. As per the former reason, for those datasets that have relatively simpler characteristic distributions such as the CelebA-HQ, both the relatively complex task of deshuffling and the lesser complex task of rotation exhibit comparable performances. However, for other datasets that comprise a higher level of complexity, we can conclude that the higher the complexity of the self-supervision task, the better its contribution to the GAN training. Hence,  it is plausible to state that the deshuffling task is generally a more beneficial self-supervision task to deploy to the GAN training than the previous ones.

On the point of the importance of the baseline architecture when the deshuffling is deployed, we observe that the deshuffling improves the performance of the SNGAN in terms of FID measure by a considerable amount on LSUN datasets as we demonstrate, while it improves the performance of the RaLSGAN on LSUN-Church and CelebA-HQ datasets (by 2.5\% for LSUN-Church, by 11\% for CelebA-HQ). We also observe that the effects of the deshuffling task on the RaLSGAN performance is minimal by small amounts as Table~\ref{tab:eval} presents. This may be because the RaLSGAN setting is already stabilized and its capacity to solve the problems is enough. We note that this stabilization and the enhanced capacity of the RaLSGAN might be due to the relativistic nature of the objectives used in the training. As Jolicoeur-Martineau \cite{ragan} states, the relativism idea enables the discriminator to actually learn from the real data unlike the classical adversarial training. Likewise, the self-supervision tasks are also beneficial for the discriminator to aid it in learning the real data distribution without forgetting it. Since the relativistic objectives cooperate with the discriminator, the contribution of a self-supervision task is limited. We further conjecture that as the architectural design for the RaLSGAN, which is based on the DCGAN, and that of the SNGAN differ, the design of the discriminator might matter such that the capacity of the discriminator to solve the additional tasks and its availability for the enhancements affect how much the self-supervision task can help towards an improved learning.

Apart from the LSUN datasets, CelebA-HQ is the only dataset that the deshuffling self-supervision seems to be not cooperative to improve the learning for the SNGAN setting. \cite{ssgan} also shows that the rotation prediction self-supervision cannot get ahead of the SNGAN. 

CelebA-HQ has a certain structural space based on the faces of humans, that is constrained in a specific way. This indicates that the complexity of the underlying data distribution is less than that of the  environments in the LSUN data spaces. While the LSUN data spaces represent a wide range of diversity in both indoor and outdoor scenes, CelebA-HQ has a comparably constrained structural space, meaning that the faces of the humans are structurally similar and the general face form does not change from human to human. Assuming that there is not a physical anomaly in the face, the parts of the human faces such as eyes, nose, or mouth are generally in the same structural positions. Along with these structural properties in the dataset, CelebA-HQ is originally pre-processed in a way that the faces are always centered in the images. We call all such characteristics of the dataset as the semantics of the dataset. When we analyze the semantics of the LSUN datasets, we can figure out the positional, textural, and structural diversity of the objects as the shapes and appearances of the objects in the bedroom scenes vary, or the facades of the church building have different architectural designs. The positions of the objects or buildings in the images of the LSUN datasets also vary unlike the CelebA-HQ dataset. As their semantics resolve the complexity of the datasets, we conjecture that the reason for both self-supervision tasks, namely the deshuffling and rotation prediction, to be not supportive of the learning features of CelebA-HQ is due to mainly the semantics of the dataset.

\begin{figure*}[t]
    \centering
    \includegraphics[width=0.99\linewidth]{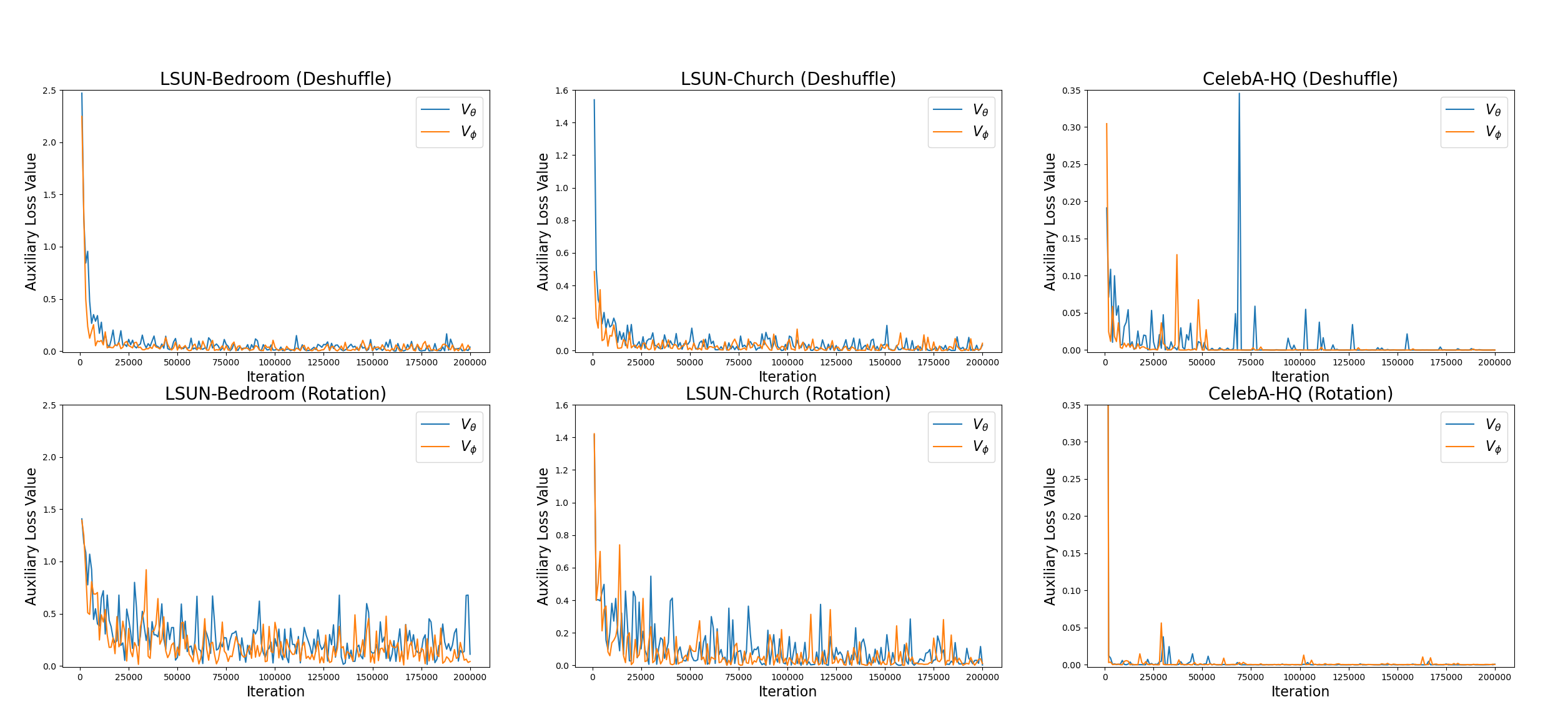}
    \caption{The deshuffler loss values of SNGAN + Deshuffle and SS-GAN settings for all datasets. Self-supervision tasks seem to be non-cooperative for the learning process over the CelebA-HQ dataset.}
    \label{fig:loss}
\end{figure*}

Apart from the semantics of the datasets, we also study the self-supervision losses over different datasets and observe that the speed for learning to deshuffle actually matters. As Fig~\ref{fig:loss} shows, the auxiliary deshuffling losses $V_\theta$ and $V_{\phi}$ of the $D$ and the $G$ respectively for both rotation prediction and deshuffling tasks on CelebA-HQ dataset start nearly at 0 and do not seem to improve. These results for both the deshuffling and rotation predictions tasks indicate that actually, these self-supervision tasks are not effective enough on the loss landscape during the training. On the other hand, the loss curves of the deshuffling task on LSUN-Bedroom and LSUN-Church datasets seem to be cooperative on learning. Compared to the rotation prediction task, the loss curves of the deshuffling for LSUN-Bedroom and LSUN-Church datasets seem more stabilized and smoother. 

The most important contribution of the self-supervision framework for the GAN training is that the self-supervised GANs are shown to alleviate the catastrophic forgetting problem \cite{ssgan}. Catastrophic forgetting problem arises from the non-stationary nature of the GAN training. In an unconditional setting, $D$ tends to forget the earlier tasks, and self-supervision is shown to help to avoid this problem in \cite{ssgan}.

When we evaluate the representation quality of the learnt features for both the conditional DeshuffleGAN and the conditional SNGAN, we also demonstrate that the enhancement of the $D$ by the deshuffling task alleviates the catastrophic forgetting in Fig.~\ref{fig:imagenet}, as the $D$ of the conditional SNGAN seems to forget the useful information after the 400K iterations while the $D$ of the conditional DeshuffleGAN seems to carry the information during the GAN training. Therefore, we can state that the DeshuffleGANs improve the training stability so that they can better address the catastrophic forgetting problem than their baseline counterparts.

Other than the generation and the representation quality, studying the deshuffling operation on its own shows the importance of the number of tiles to use, and the performance of the discriminator in solving a jigsaw puzzle. The first thing that comes to mind as a reason for why 9 tiles yield a better performance is the complexity of the task since deshuffling 9 tiles is more complicated than deshuffling 4 tiles. Another reason might be due to the distances among the set of permutations, as the permutation list for the 9 tiles is formed by the maximum Hamming distances, while the permutation list for the 4 tiles consists of all the possible $4!$ permutations. The latter permutations involve  closer distances to each other.

\begin{table}[t]
  \begin{center}
    \caption{Deshuffling accuracies of different discriminators.}
    \label{tab:accuracy}
    \begin{tabular}{ccc} 
        \hline
        TRAIN DATASET &TEST DATASET &ACCURACY (\%) \\
        \hline
        \multirow{4}{*}{LSUN-Church} & LSUN-Church (real) &98.85 \\
        & LSUN-Church (fake) &99.50\\
        & LSUN-Bedroom (real) &73.50\\
        & LSUN-Bedroom (fake) &74.06\\
        \hline
        \multirow{4}{*}{LSUN-Bedroom} & LSUN-Bedroom (real) &98.39 \\
        & LSUN-Bedroom (fake) &99.09\\
        & LSUN-Church (real) &86.97\\
        & LSUN-Church (fake) &86.93\\
        \hline
    \end{tabular}
  \end{center}
\end{table}

The deshuffling performances of different discriminators indicate that the discriminators are capable of solving the jigsaw puzzles. When we monitor the accuracies, we draw a conclusion that the generator tries to synthesize structurally coherent images in a way that their jigsaw puzzles can be deshuffled by the discriminator. That is why, the accuracies for the fake samples are even higher than the real samples (Table~\ref{tab:accuracy}). Aside from testing on the same dataset with which the discriminator is trained, we also use real and fake samples from different datasets in the test phase to view the generalizability of the learnt features to solve the deshuffling task. We notice that the discriminators trained with the LSUN-Bedroom dataset perform better in deshuffling the samples from the LSUN-Church dataset than the opposite case. The justification for these results may be due to the characteristics of the datasets such that even if the number of samples used in the GAN training are the same because of the same iteration numbers and the batch sizes, when the LSUN-Bedroom dataset is used in the training, batches are sampled from approximately 3M samples capturing diverse bedroom scenes, while LSUN-Church comprises of only 126K samples, hence it is less varied in nature than the former. As the possibility of sampling images containing diverse features is higher in the LSUN-Bedroom dataset, the features that are learned in order to deshuffle the images capture more generalizable characteristics.

\section{Conclusion}

In this work, we present the DeshuffleGAN and the cDeshuffleGAN that incorporate the deshuffling task, which is previously used only as a pretext in self-supervised learning, into the GAN framework. This incorporation increases the discriminator's representation power through improved structure learning. 

Compared to the baseline versions, the visual inspection shows particularly that a more consistent generation quality is obtained by DeshuffleGANs because they struggle to put the pieces of an unshuffled image together while trying to generate realistic outputs, and this pushes them to learn improved features to represent the global spatial structure and appearance of the input data space as well as its local structural continuity and coherence.

We show that DeshuffleGANs can improve the performance of different GAN architectures which indicate the generalizability of the deshuffling task. We also use various datasets for the generations in order to prove the capability of DeshuffleGANs over different data distributions. DeshuffleGANs can manage to generate images with complex representations such as buildings, human faces, or bedrooms. However, we show that the performance of the DeshuffleGAN varies with respect to the complexity of the underlying data distribution. Within more complex environments such as the LSUN data spaces, the DeshuffleGAN demonstrates its well-enhancing effects, whereas, for more constrained structural spaces such as the faces of humans, its effect is inconsequential.

Furthermore, we compare the deshuffling task with the rotation prediction task. We show that for most of the settings and via FID curves, the deshuffling task aids the adversarial training more than the rotation task. We hypothesize that the reason is due to the relatively higher complexity of the deshuffling task compared to that of the rotation prediction task. As the discriminator learns to solve a more complex structural problem in deshuffling, the discriminator gets further enhanced compared to the rotation prediction, which is only a global transformation problem. We also demonstrate that the selected number of tiles in the deshuffling process matters; as the search space is enlarged, the prediction task is complexified, further strengthening the self-supervision signal.

As future work, the reasons why the deshuffling task does not improve the baseline performance on some datasets can be analyzed deeply. In this work, we show the non-cooperative behavior of the auxiliary tasks for CelebA-HQ dataset on the loss landscape. However, there remains open questions on what kind of and how auxiliary signals should be incorporated within various settings.

\section*{Acknowledgment}
In this work, Gulcin Baykal and Furkan Ozcelik were supported by the Turkcell-ITU Researcher Funding Program. Gulcin Baykal was further supported by DeepMind Scholarship Program at ITU. This work was also supported by the Scientific Research Project Unit of Istanbul Technical University [project number MOA-2019-42321].

\bibliographystyle{elsarticle-num}
\bibliography{mybibfile}

\begin{thebibliography}{10}
\expandafter\ifx\csname url\endcsname\relax
  \def\url#1{\texttt{#1}}\fi
\expandafter\ifx\csname urlprefix\endcsname\relax\def\urlprefix{URL }\fi
\expandafter\ifx\csname href\endcsname\relax
  \def\href#1#2{#2} \def\path#1{#1}\fi

\bibitem{gan}
I.~Goodfellow, J.~Pouget-Abadie, M.~Mirza, B.~Xu, D.~Warde-Farley, S.~Ozair,
  A.~Courville, Y.~Bengio, Generative adversarial nets, in: Advances in Neural
  Information Processing Systems, 2014.

\bibitem{autoencoder}
H.~Bourlard, Y.~Kamp, Auto-association by multilayer perceptrons and singular
  value decomposition, Biol. Cybern. 59~(4–5) (1988) 291–294.

\bibitem{gansurvey}
Z.~Wang, Q.~She, T.~E. Ward, Generative adversarial networks: A survey and
  taxonomy, ArXiv abs/1906.01529.

\bibitem{ssgan}
T.~Chen, X.~Zhai, M.~Ritter, M.~Lucic, N.~Houlsby, Self-supervised gans via
  auxiliary rotation loss, in: 2019 IEEE/CVF Conference on Computer Vision and
  Pattern Recognition (CVPR), 2019, pp. 12154--12163.
\newblock \href {http://dx.doi.org/10.1109/CVPR.2019.01243}
  {\path{doi:10.1109/CVPR.2019.01243}}.

\bibitem{ssgan_minimax}
N.-T. Tran, V.-H. Tran, N.-B. Nguyen, L.~Yang, N.-M. Cheung, Self-supervised
  gan: Analysis and improvement with multi-class minimax game, in: Advances in
  Neural Information Processing Systems, 2019.

\bibitem{fxgan}
R.~Huang, W.~Xu, T.-Y. Lee, A.~Cherian, Y.~Wang, T.~K. Marks, Fx-gan:
  Self-supervised gan learning via feature exchange, 2020 IEEE Winter
  Conference on Applications of Computer Vision (WACV) (2020) 3183--3191.

\bibitem{deshufflegan}
G.~{Baykal}, G.~{Unal}, Deshufflegan: A self-supervised gan to improve
  structure learning, in: 2020 IEEE International Conference on Image
  Processing (ICIP), 2020, pp. 708--712.

\bibitem{catastrophic}
J.~Kirkpatrick, R.~Pascanu, N.~Rabinowitz, J.~Veness, G.~Desjardins, A.~A.
  Rusu, K.~Milan, J.~Quan, T.~Ramalho, A.~Grabska-Barwinska, et~al., Overcoming
  catastrophic forgetting in neural networks, Proceedings of the national
  academy of sciences 114~(13) (2017) 3521--3526.

\bibitem{lsun}
F.~Yu, Y.~Zhang, S.~Song, A.~Seff, J.~Xiao, Lsun: Construction of a large-scale
  image dataset using deep learning with humans in the loop, arXiv preprint
  arXiv:1506.03365.

\bibitem{progan}
T.~Karras, T.~Aila, S.~Laine, J.~Lehtinen, Progressive growing of {GAN}s for
  improved quality, stability, and variation, in: International Conference on
  Learning Representations, 2018.

\bibitem{spectral}
T.~Miyato, T.~Kataoka, M.~Koyama, Y.~Yoshida, Spectral normalization for
  generative adversarial networks, in: International Conference on Learning
  Representations, 2018.

\bibitem{dcgan}
A.~Radford, L.~Metz, S.~Chintala, Unsupervised representation learning with
  deep convolutional generative adversarial networks (2015).
\newblock \href {http://arxiv.org/abs/1511.06434} {\path{arXiv:1511.06434}}.

\bibitem{ragan}
A.~Jolicoeur-Martineau, The relativistic discriminator: a key element missing
  from standard gan, in: International Conference on Learning Representations,
  2019.

\bibitem{imagenet}
J.~Deng, W.~Dong, R.~Socher, L.-J. Li, K.~Li, L.~Fei-Fei, Imagenet: A
  large-scale hierarchical image database, in: 2009 IEEE Conference on Computer
  Vision and Pattern Recognition (CVPR), 2009, pp. 248--255.
\newblock \href {http://dx.doi.org/10.1109/CVPR.2009.5206848}
  {\path{doi:10.1109/CVPR.2009.5206848}}.

\bibitem{fid}
M.~Heusel, H.~Ramsauer, T.~Unterthiner, B.~Nessler, S.~Hochreiter, Gans trained
  by a two time-scale update rule converge to a local nash equilibrium, in:
  Advances in Neural Information Processing Systems, 2017, p. 6629–6640.

\bibitem{equal_gan}
M.~Lucic, K.~Kurach, M.~Michalski, O.~Bousquet, S.~Gelly, Are gans created
  equal? a large-scale study, in: Advances in Neural Information Processing
  Systems, 2018, p. 698–707.

\bibitem{colorization}
R.~Zhang, P.~Isola, A.~A. Efros, Colorful image colorization, in: ECCV, 2016.

\bibitem{gp}
I.~Gulrajani, F.~Ahmed, M.~Arjovsky, V.~Dumoulin, A.~C. Courville, Improved
  training of wasserstein gans, in: Advances in Neural Information Processing
  Systems, 2017, pp. 5767--5777.

\bibitem{spectral_bounding}
Z.~Zhang, Y.~Zeng, L.~Bai, Y.~Hu, M.~Wu, S.~Wang, E.~R. Hancock, Spectral
  bounding: Strictly satisfying the 1-lipschitz property for generative
  adversarial networks, Pattern Recognition 105 (2020) 107179.

\bibitem{wasserstein}
M.~Arjovsky, S.~Chintala, L.~Bottou, Wasserstein gan, in: International
  Conference on Machine Learning, 2017, pp. 214--223.

\bibitem{lsgan}
X.~Mao, Q.~Li, H.~Xie, R.~Y.~K. Lau, Z.~Wang, S.~P. Smolley, Least squares
  generative adversarial networks, 2017 IEEE International Conference on
  Computer Vision (ICCV) (2017) 2813--2821.

\bibitem{tackle}
W.~Li, L.~Fan, Z.~Wang, C.~Ma, X.~Cui, Tackling mode collapse in
  multi-generator gans with orthogonal vectors, Pattern Recognition 110 (2021)
  107646.
\newblock \href
  {http://dx.doi.org/https://doi.org/10.1016/j.patcog.2020.107646}
  {\path{doi:https://doi.org/10.1016/j.patcog.2020.107646}}.

\bibitem{cgan}
M.~Mirza, S.~Osindero, Conditional generative adversarial nets, ArXiv
  abs/1411.1784.

\bibitem{projection}
T.~Miyato, M.~Koyama, c{GAN}s with projection discriminator, in: International
  Conference on Learning Representations, 2018.

\bibitem{infogan}
X.~Chen, Y.~Duan, R.~Houthooft, J.~Schulman, I.~Sutskever, P.~Abbeel, Infogan:
  Interpretable representation learning by information maximizing generative
  adversarial nets, in: Advances in Neural Information Processing Systems,
  2016.

\bibitem{acgan}
A.~Odena, C.~Olah, J.~Shlens, Conditional image synthesis with auxiliary
  classifier gans, in: International Conference on Machine Learning, 2017.

\bibitem{sagan}
H.~Zhang, I.~J. Goodfellow, D.~N. Metaxas, A.~Odena, Self-attention generative
  adversarial networks, in: International Conference on Machine Learning, 2019,
  pp. 7354--7363.

\bibitem{biggan}
A.~Brock, J.~Donahue, K.~Simonyan, Large scale {GAN} training for high fidelity
  natural image synthesis, in: International Conference on Learning
  Representations, 2019.

\bibitem{sssurvey}
L.~Jing, Y.~Tian, Self-supervised visual feature learning with deep neural
  networks: A survey, IEEE transactions on pattern analysis and machine
  intelligence.

\bibitem{laplacian}
Q.~Wang, H.~Fan, G.~Sun, Y.~Cong, Y.~Tang, Laplacian pyramid adversarial
  network for face completion, Pattern Recognition 88 (2019) 493--505.
\newblock \href
  {http://dx.doi.org/https://doi.org/10.1016/j.patcog.2018.11.020}
  {\path{doi:https://doi.org/10.1016/j.patcog.2018.11.020}}.

\bibitem{jigsaw}
M.~Noroozi, P.~Favaro, Unsupervised learning of visual representations by
  solving jigsaw puzzles, in: ECCV, 2016.

\bibitem{rotation}
S.~Gidaris, P.~Singh, N.~Komodakis, Unsupervised representation learning by
  predicting image rotations, in: International Conference on Learning
  Representations, 2018.

\bibitem{context}
C.~Doersch, A.~Gupta, A.~A. Efros, Unsupervised visual representation learning
  by context prediction, 2015 IEEE International Conference on Computer Vision
  (ICCV) (2015) 1422--1430.

\bibitem{revisitingss}
A.~Kolesnikov, X.~Zhai, L.~Beyer, Revisiting self-supervised visual
  representation learning, 2019 IEEE/CVF Conference on Computer Vision and
  Pattern Recognition (CVPR) (2019) 1920--1929.

\bibitem{multitaskss}
C.~Doersch, A.~Zisserman, Multi-task self-supervised visual learning, 2017 IEEE
  International Conference on Computer Vision (ICCV) (2017) 2070--2079.

\bibitem{multidomainss}
Z.~Feng, C.~Xu, D.~Tao, Self-supervised representation learning from
  multi-domain data, 2019 IEEE/CVF International Conference on Computer Vision
  (ICCV) (2019) 3244--3254.

\bibitem{jigen}
F.~M. Carlucci, A.~D'Innocente, S.~Bucci, B.~Caputo, T.~Tommasi, Domain
  generalization by solving jigsaw puzzles, in: 2019 IEEE/CVF Conference on
  Computer Vision and Pattern Recognition (CVPR), 2019, pp. 2229--2238.

\bibitem{channel_info}
K.~Kurach, M.~Lu{\v{c}}i{\'c}, X.~Zhai, M.~Michalski, S.~Gelly, A large-scale
  study on regularization and normalization in {GAN}s, Vol.~97 of Proceedings
  of Machine Learning Research, PMLR, 2019, pp. 3581--3590.

\bibitem{adam}
D.~P. Kingma, J.~Ba, Adam: A method for stochastic optimization, in:
  International Conference on Learning Representations, 2015.

\end{thebibliography}

\end{document}